\documentclass[lettersize,journal]{IEEEtran}
\usepackage{amsmath,amsfonts}
\usepackage{algorithmic}
\usepackage{algorithm}
\usepackage{array}
\usepackage[caption=false,font=normalsize,labelfont=sf,textfont=sf]{subfig}
\usepackage{textcomp}
\usepackage{stfloats}
\usepackage{url}
\usepackage{verbatim}
\usepackage{graphicx}
\usepackage{cite}
\begin{document}

\title{On the Road to More Accurate \\ Mobile Cellular Traffic Predictions}

\author{Natalia Vassileva Vesselinova \\ Department of Mathematics and Systems Analysis \\ Aalto University School of Science
	% <-this % stops a space
	\thanks{The ideas and findings reported in this paper were produced during 2021 while the author was contributing to the 5G-ROUTES project. The main concept and its application to network slicing in 5G for CAM services was filed for patenting at the end of December 2022.}% <-this % stops a space
	\thanks{This work was supported by the European Commission under the 5G-ROUTES (GA 951867) project. }}

% The paper headers
%\markboth{Journal of \LaTeX\ Class Files,~Vol.~14, No.~8, April~2023}%
%{Shell \MakeLowercase{\textit{et al.}}: A Sample Article Using IEEEtran.cls for IEEE Journals}

%\IEEEpubid{0000--0000/00\$00.00~\copyright~2021 IEEE}
% Remember, if you use this you must call \IEEEpubidadjcol in the second
% column for its text to clear the IEEEpubid mark.

\maketitle

\begin{abstract}
We introduce a conceptually novel approach 
to forecasting mobile cellular traffic. 
It is especially relevant to 
the future wireless communication networks 
that will support automated driving 
together with various innovative %time-sensitive, 
connected and automated mobility services. 
%with time bounds. 
Our focus is on the less extensively studied 
highway scenario, % and on cross borders,
where the load on the mobile cellular system 
can change drastically during short time intervals 
due to the highly dynamic vehicular flow.
%of time 
%and the network resources are orchestrated proactively.
We employ a recurrent neural network 
suitable for learning from time series data 
to address the challenging case of 
short-term predictions at a cell level.
The prediction error is decreased
by exploiting metrics that account for 
the data generation function and spatial dependences.
The improvement in prediction performance  
%attained 
under the examined conditions
is from $22.4\%$ up to $68\%$ % for the evaluated conditions
over the generalised traditional approach, 
which uses purely network metrics.
\end{abstract}


\section{Introduction}
\IEEEPARstart{T}{raffic} prediction has  been an integral part of 
the management of the mobile cellular networks ever since their commercial use. 
As the fifth generation (5G) and beyond 5G communication systems 
are adopting dynamic resource orchestration 
to accommodate the diverse needs of various industries 
with different service requirements over the same physical infrastructure,
forecasting becomes central for their operation,
especially when delivering time-critical and safety-critical applications.

Traditionally, the prediction of the mobile cellular traffic 
has been based solely on its historical measurements. 
More recently, 
with the adoption of machine learning in the mobile networks,
other cellular key performance indicators (KPIs) 
have been harnessed \cite{piRoad}. 
In a recent study \cite{ho}, %for instance, 
the handover frequency is exploited
as a means to model the mobility of the users
and hence, better predict the traffic load
in base stations (BSs). 
In other works, such as \cite{crossDomain}, 
meta information is used to
capture external factors that might have an impact on the 
traffic patterns. Often, the augmented data sets 
include the location of points of interest 
(commercial centres, touristic spots or sport venues to name a few)
as they exhibit very distinct traffic dynamics. 
In the latest state-of-the-art research, 
such as \cite{piRoad, ho, crossDomain, stTran, eff, agg, IoT}, 
the focus is on developing 
advanced learning structures that can model
the non-linearities between the traffic observed in 
different BSs (locations) of the mobile network.

We propose a paradigm shift by turning our attention to the data.
It stems from the realisation that in contrast to the past,
nowadays we have the means to count the number of 
potential users of the network and accordingly, of its services. 
Such metric characterizes the data generation process.
Thereby, this statistic has the potential 
to decrease the uncertainty 
and increase the prediction accuracy.
In our research we focus on highways and vehicle-to-everything 
services with time bounds.
We count the number of vehicles on the highway 
and measure their average velocity.
The incorporation of these road metrics 
is motivated by the observation that 
V2X traffic on a highway
can be generated by 
the vehicles on that highway.

\begin{figure}[tp]
	\centering
	\includegraphics[width=\columnwidth]{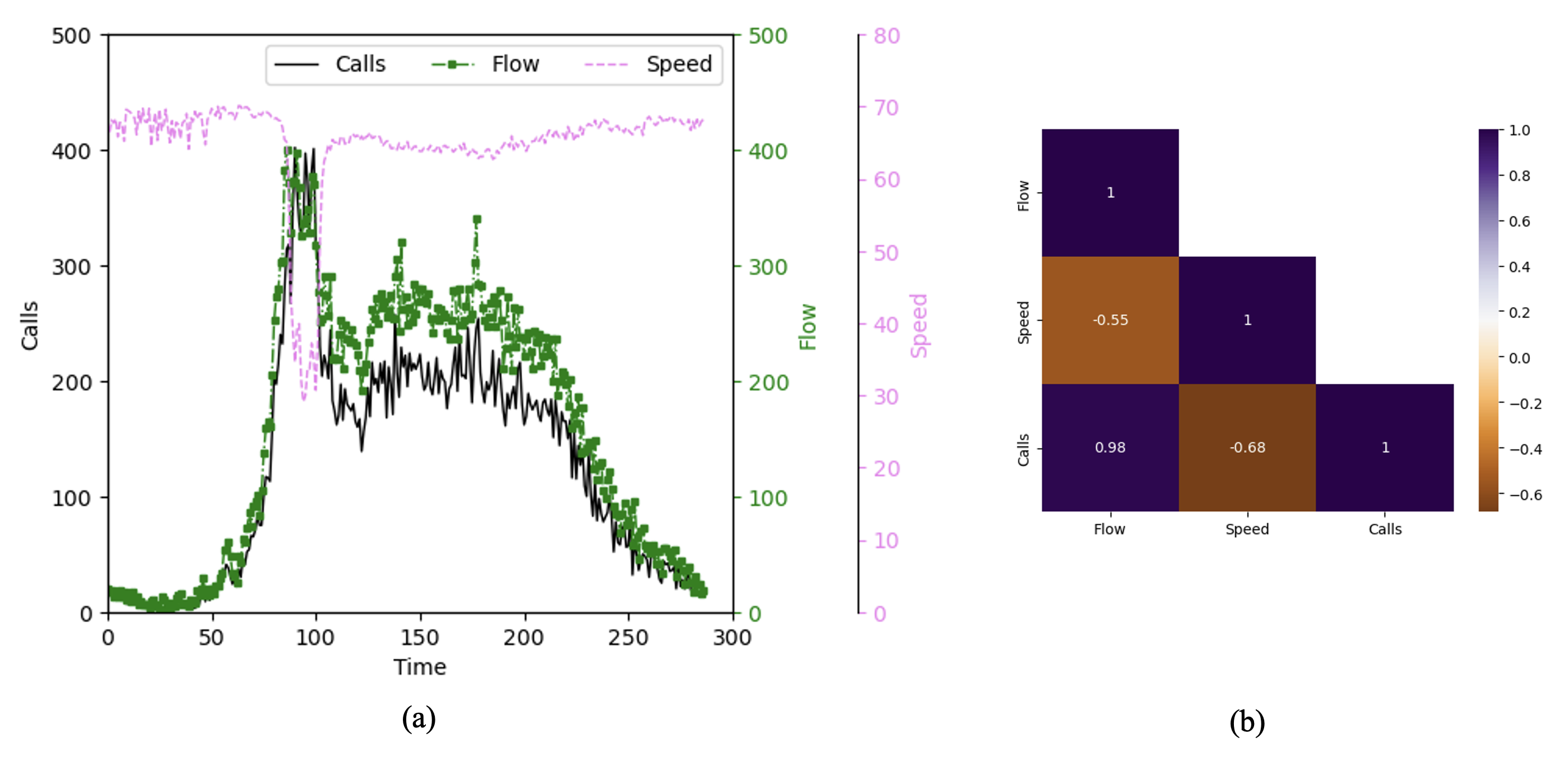}
	\caption{(a) PeMS road flow and speed measurements for a single work day and calls generated according to Poisson process with rate $\lambda = 1/5$ and (b)~a heat map correlation matrix of the three variables.}
	\label{fig:corr}
\end{figure}

Heterogeneous data sets have previously been used 
by the machine learning community.
As an example, weather---especially severe conditions---has 
a measurable effect on the traffic flow and speed.
In effect, it has been shown that 
machine learning 
prediction accuracy is increased by 
incorporating relevant weather features
into the data.
Our approach conceptually differs from such studies
as they employ meta data to account for external factors 
that might impact the estimated variable.
Instead of auxiliary information, we employ metrics 
of the data generation process.

Road traffic forecasting has earlier been introduced  
\cite{autoMEC} 
to support the timely orchestration of 
multi-access edge computing (MEC) resources.
A decision method in \cite{autoMEC}  is fed with  
road flow predictions
in order to manage MEC resources 
and meet service requirements.
The number of service requests is obtained 
in \cite{autoMEC} 
by mapping road traffic predictions 
to network modeling assumptions.
Among other, it is assumed that every vehicle is involved 
in a single connection, which opposes 
the understanding that in the future 
a single vehicle can be a source of 
multiple simultaneous communication sessions 
(for autonomous driving, infotainment and maintenance for instance) 
\cite{campolo}. 
In the study reported below, 
we do not forecast road traffic flow nor
make assumptions about the number of 
communication sessions 
each vehicle generates 
but model the latter with a random process.
We use machine learning to forecast the mobile traffic load on the BS. 
\\
\\
Finally, due to the very dynamic vehicular and hence network flow,
our interest is in short-term forecasting.
It is more challenging to predict over short than long time horizons 
because the traffic can quickly change volume and direction 
within minutes %(see Fig. \ref{fig:corr}), 
whereas over hours and days, %the traffic is smoother and 
it is smoother and
exhibits clear periodicity \cite{IoT, bigData}.

 \section{Problem formulation}

Let $\mathbf{x}^\tau \in R^p$ 
denote a multivariate random vector 
of measurements of metrics
relevant to a cell.
Given a time sequence of $M$ such historical observations 
$\{ \mathbf{x}^{\tau - M + 1}, \dots, \mathbf{x}^\tau \}$,  
our objective is to learn a prediction model $\mathcal{F}$ 
that can forecast the future mobile traffic load
$\hat{x}$ in the cell during the next $T$ time step(s):

\begin{equation}
	\hat{x}^{\tau + 1}, \dots , \hat{x}^{\tau + T} = \mathcal{F} (\mathbf{x}^{\tau - M + 1}, \dots, \mathbf{x}^\tau). \label{focus}
\end{equation}
so that the prediction error is mininized:
\begin{equation*}
	\text{min} \: L(\hat{x}, x),
\end{equation*}
where the loss function $L(\mathord{\cdot})$ measures the difference between 
the estimated $\hat{x}$ and observed $x$ mobile cellular traffic. 

Without loss of generality, 
the predicted variable is 
the number of service requests
or equivalently---under sufficient capacity---the 
calls served  by a base station. 
This is in line with the state-of-the-art  works 
\cite{crossDomain, stTran, eff, agg, citywide}
that measure cellular traffic by 
the number of served calls. 
Note that the principles of 
forecasting V2X cellular traffic 
we propose in this study 
can be applied together with 
any suitable traffic metric or KPIs
and integrated in practically any learning approach.

\begin{figure}[!t]  %[htbp]
	\includegraphics[width=\columnwidth]{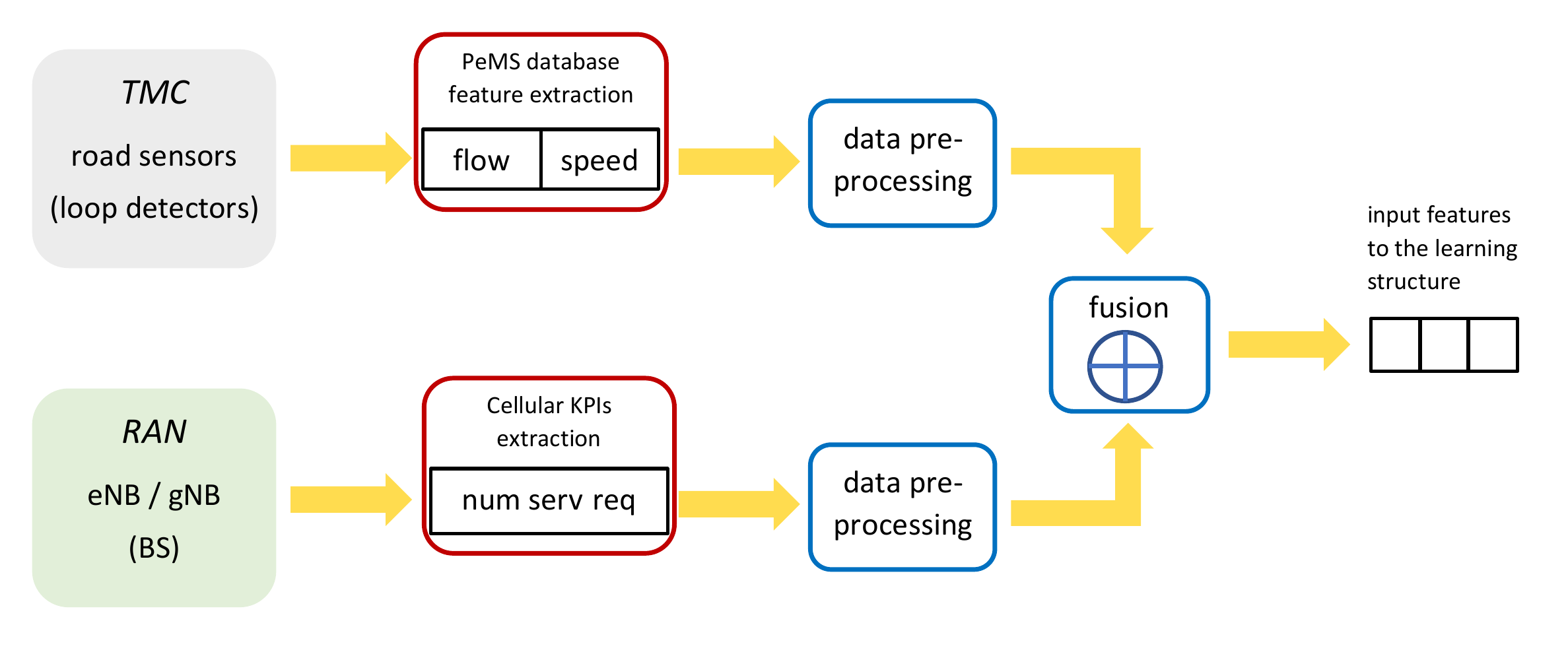}   %0.5\textwidth
	\caption{Data are collected by the traffic monitoring system (TMS) and radio access network (RAN) and preprocessed. The output is a vector obtained by concatenating road with cellular traffic feaures.}
	\label{fig:data}
\end{figure}

\section{Data}
\label{sec:data}

\subsection{Choice of metrics}

Measuring and recording KPIs is 
an intrinsic part of any mobile cellular system 
as the measurements are used for 
monitoring, operating, maintaining and optimising the network.
Likewise, road traffic measurements 
are essential to any off-the-shelf 
traffic management system (TMS).
In contrast to mobile systems, 
traffic measurements are open access. 
The California Department of Transportations 
(Caltrans), for instance, 
is maintaining a Performance Measurement System 
(PeMS) \cite{pems}, 
which collects real time traffic data from 
a large number of individual detectors
deployed statewide in the California freeway system.
Both real-time and historical measurements are freely 
available via the PeMS database.
Such TMSs exist across major cities and highways 
in Europe, Asia and in the Americas. 

The setting for collecting road and cellular  traffic 
in real-life systems is illustrated in Fig.~\ref{fig:data}.
The road metrics employed are vehicular flow 
(the number of vehicles that pass by a specific location 
during a time interval) and the average speed of the flow. 
In the PeMS data set, 
the former metric is aggregated  
and the latter averaged 
over five-minute intervals.

A heatmap of the correlation between the flow 
and the speed, Fig. \ref{fig:corr}, shows 
that they are inversely (negatively) correlated.
Despite that they are correlated mainly 
during peak hours, we experimentally confirmed 
our hypothesis that the prediction accuracy 
is increased  when speed measurements 
are  incorporated into the data.
The speed metric 
can be an (early) indicator of sudden changes 
in the vehicular flow such as traffic jams or incidents for instance, Fig. \ref{fig:corr}.

\begin{figure}[!t]	
	\centering
	\includegraphics[width=2.75in]{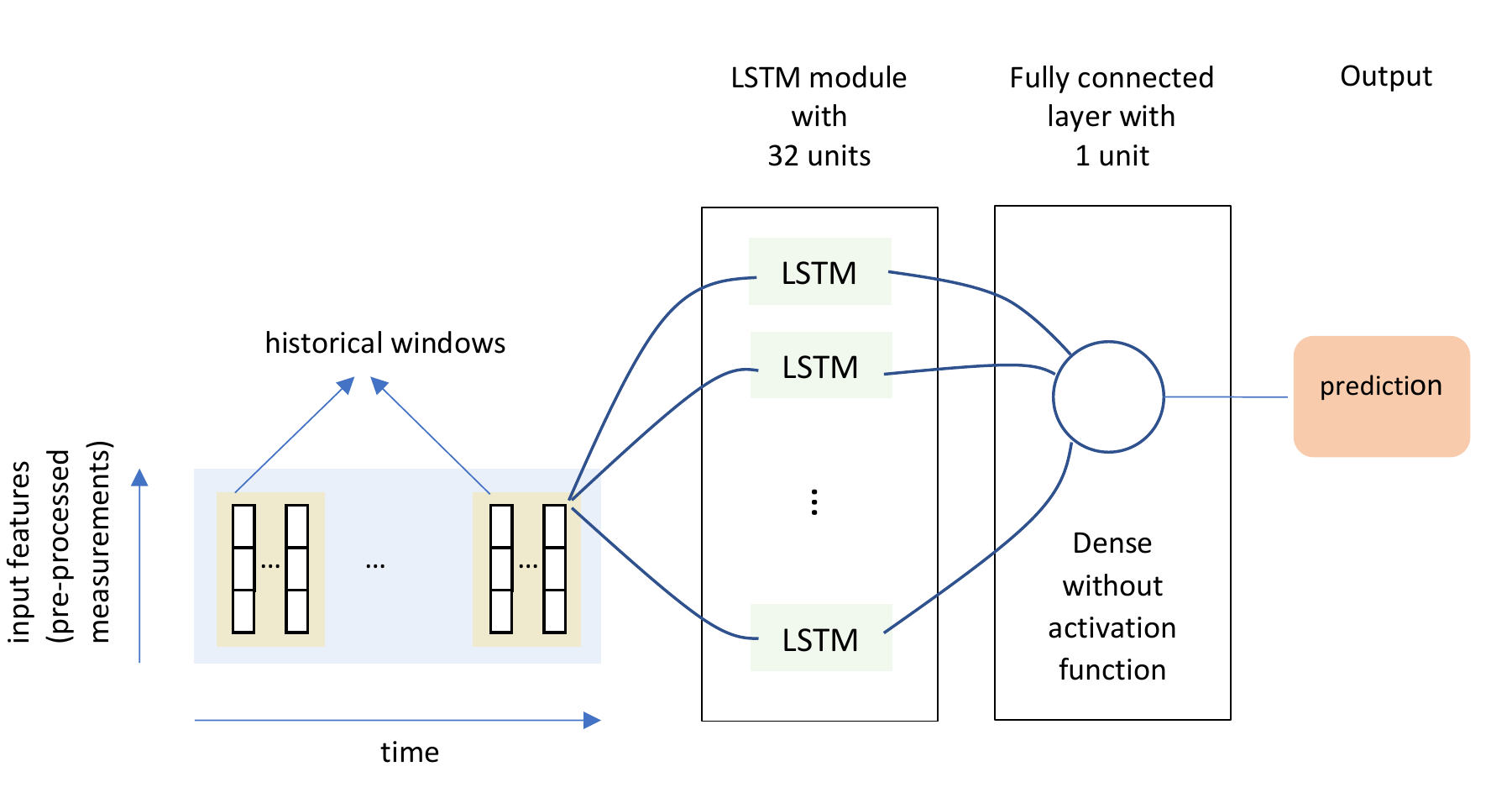} %
	\caption{The learning structure is based on an LSTM module and a fully connected feed-forward neural network. The model is fed with the preprossed and fused road with cellular traffic samples.}
	\label{fig:learningmodel}
\end{figure}

\subsection{Description}

\textbf{Road data.} Real-life road traffic data 
collected from a unidirectional, three-lane freeway, 
in the outskirts of San Francisco is used in the experiments \cite{pems}.
The data comprise road metrics recorded for each work day of a week, 
from Monday to Friday,  from week 13 (March 29, 2021) until week 32 (August 13, 2021). 
There are 288 data points per day due to the 5-minute granularity.

We make an assumption that road metrics are collected outside yet in the vicinity of the BS, 
so that their measurements in a prior observation interval %(during time step $\tau$ for instance) 
are relevant to the time interval for which cellular traffic predictions are made. %($\tau+1$).

\textbf{Mobile cellular data.} Since there are no open access data
that combine mobile cellular with highway traffic measurements, 
we generate the former based on the PeMS real-life road data and the 
modeling assumptions described next.

Service requests are generated by each vehicle
with a common rate $\lambda$ 
following a Poisson process.
As this is a random process, 
each passing vehicle can generate no service request, 
at most one, or multiple requests 
while driving through the segment of the freeway 
covered by the BS.
The service generation processes is preceded by emulation of 
exponentially distributed inter-arrival times
of the vehicular flow measured by PeMS 
during each five-minute  interval.

\subsection{Preprocessing}
Small deviations from the average speed 
observed during prolonged periods of time do not have important 
effect on the vehicular flow, vehicle's dwell time in the cell
and hence on the number of service requests.
Therefore, we discretize the speed into 8 levels to ease the learning process. 
Speeds below 20 mph are labelled by 1, and 
successively by integers up to 8 for speeds
above 60 mph.
Our experimental results show that 
the prediction error can be decreased by roughly $2 \%$
just by applying this simple pre-processing step.

All data features---vehicular flow, speed level and number of service requests---are
normalised by the mean and standard deviation of the training data,
individually for each feature, so that they all
take on the same scale.

After this pre-processing step the road and cellular traffic features are fused
by using a  concatenation operation, Fig. \ref{fig:data}. The resultant  vector $\mathbf{x} \in R^3$ is 
a single sample from a historical window, which the machine learning model 
described next is fed with.

\section{Learning model}
\label{sec:model}

The cellular traffic prediction model is usually learnt by accounting for 
temporal correlations alone or 
by considering spatial correlations between different BSs  (locations) too. 
Temporal correlations are normally learnt with recurrent neural networks
as these are designed to extract patterns from sequential data.
Similarly, spatial correlations are captured with applicable learning structures
such as graph neural networks. 

%We propose to train a machine learning model with road traffic and cellular traffic metrics.
We focus on learning a short-term forecasting model at a BS level as formulated in (\ref{focus}). 
Since during short time horizons, only traffic in the cell and its immediate vicinity is 
relevant to traffic prediction, the spatial correlations are incorporated in the dataset
via the flow and speed, which are measured in the vicinity yet outside the BS, 
instead of explicitly through the learning structure. 
To our knowledge, similar approach has not been explored before. 

The structure of the learning model is illustrated in Fig.~\ref{fig:learningmodel}.
It consists of a long short term memory (LSTM)~\cite{LSTM}, 
which has been designed to overcome 
the vanishing gradient problem typical for 
traditional recurrent neural networks.
The LSTM module is followed by 
a fully connected feedforward neural network 
(a multilayer perceptron).
This dense layer consists of a single unit  
and no activation function. 
It is a linear transformation that maps 
the final LSTM state
into a single target value --  
the estimated number of calls at the BS.

Mathematically, the model can be represented 
as two functions $f(g),$ where
$g(\cdot)$ is the LSTM learning structure,
which transforms the input  data into new features.
The representation function $f(\cdot)$ 
maps the learnt features into 
a cellular traffic prediction for the BS under consideration.

All results reported below are obtained with a model 
composed by a single LSTM layer of 32 cells.
We obtained practically identical results with 
gated recurrent units (GRU) layer of 32 cells.  
GRU is a lighter model that can be used instead of LSTM 
when computational resources are limited. 
The RMS Prop optimiser is used in training.
The loss function used in the training phase is the mean squared error (MSE):

\begin{equation}
	\text{MSE} = \frac{1}{n} \sum_{i=1}^{n} (x_i - \hat{x}_i)^2,
\end{equation} 
where $x$ denotes the true (observed) traffic and
$\hat{x}$ is its predicted value. 

Forecasting performance is measured with 
the mean absolute error
\begin{equation}
	\text{MAE}  = \frac{1}{n} \sum_{i=1}^{n} \mid x_i - \hat{x}_i \mid.
\end{equation} 

This model can be implemented at a BS %level 
or MEC level, or in federated learning. 
The concept of using the number of cellular users 
can be integrated with centralised learning too.

\section{Results}

\subsection{Examined scenarios}

As there are no data sets that combine 
mobile cellular with road KPIs,
we create different scenarios by varying: 
the mean call arrival rate $\lambda$  
described in Section~\ref{sec:data}, 
the handover probability $h$, and by defining the range of the BS.
The mainstream 5G networks of today are deployed 
at 2.5 to 4.9~GHz (mid-band TDD) and cover roughly 
between 0.3~miles and 3~miles. 
5G coverage for wider areas  extends from 
3 to 12~miles at 0.6 -- 2.6~GHz (low-band FDD). 
Therefore,  we evaluate two deployment cases.
The other parameters, such as the time interval $\Delta=5$~min,
the vehicular flow and the average speed,
are extracted from the PeMS data described in Section~\ref{sec:data}.
The examined conditions are listed in Table~\ref{tab:results}.

The results reported below are for observation horizons of 1~hour and 30~min
($M=18$ time steps).  
The prediction horizon is the next five-minute period  
($T=1$, one time step ahead of the prediction moment).
We obtained very similar results  under 
shorter observation intervals of 30~min (6 samples).
Extending the historical window beyond 18~samples 
did not lead to any measurable improvement. 
The abrupt changes occuring in the number of calls
are due to the underlying processes---sudden 
changes in the vehicular flow and 
the randomness of the call arrival process. %, Fig. \ref{fig:corr}.
Thereby, longer historical time records 
do not seem to aid the learning process.

The first 12 weeks of data were used for training the model,
the next 4 for validating, and the last 4~weeks for testing purposes
(that is, the data was split into 3:1:1 ratio).

\begin{table}
	\caption{Scenarios and Results}
	\begin{center}
		\footnotesize{
			\begin{tabular}{| *{6}{c|}  }%{ | l | c | c | r |} 
				\hline
				\multicolumn{2}{|c|}{Radio coverage}	& $\lambda$ 	& $h$ 				& \multicolumn{2}{c|}{\textbf{MAE}} \\
				\cline{1-2} \cline{5-6}
				GH				&		miles								  & service rate 	& HO prob		& Net			& Net \& Road	\\
				\hline  
				&												 &							& 1					   & 0.1220		  & \textbf{0.0390}		   \\
				\cline{4-6}
				&												 &							& 0.8				 & 0.1430		& \textbf{0.0547}		\\
				\cline{4-6}
				2.5--4.9	 &		1.5					&	1/$\Delta$		& 0.5				 & 0.1564		& \textbf{0.0832}		  	\\
				\cline{4-6}
				&		 										 &		 					& 0.2				 & 0.1884		& \textbf{0.1210}		 \\
				\cline{4-6}
				&												 &							& 0					   & 0.1997		  & \textbf{0.1550}		   \\
				\cline{3-6}
				&												 &	 3/$\Delta$		 & 0.5				 & 0.1578		& \textbf{0.0782}		 \\
				\hline
				0.6 -- 2.6	 &		6										 &	1/$\Delta$		& 0.5				 & 0.1182		 & \textbf{0.0667}		 \\					
				\hline							
		\end{tabular}}
		\label{tab:results}
	\end{center}
\end{table}

\subsection{ Discussion} 

We evaluate the model  under  two approaches
for all examined scenarios. % case studies.
The case when the input data comprises only cellular traffic 
(number of calls)  
represents a generalised version of the conventional, 
network-centric approach \cite{bibid}, 
with KPIs that are exclusively mobile cellular.  
This case is denoted by \textit{Net} in Table~\ref{tab:results}.
We compare it with the forecasting results achieved by the same model
when trained and tested on the heterogeneous data %which includes 
through which spatial correlations 
and the underlying processes are taken into account. 
This approach is denoted by \textit{Net \& Road} 
(for network and road data) 
in Table~\ref{tab:results}.

The mean absolute prediction error is smaller when road metrics 
that  model spatial correlation and account for the data generation function
are employed and incorporated in the cellular data set.
%are employed together with a cellular traffic metric.
This is true under all studied conditions: different BS range, 
varying call arrival rate, and  handover probability.
Specifically, the test MAE for the  \textit{Net \& Road}  
is $1.3$ times lower and up to more than $3$ times smaller
than the test MAE for the  \textit{Net}.
In other words, the performance improvement 
attained by the proposed approach
under the evaluated scenarios 
is at least $22.4\%$ and as much as $68\%$ 
over the generalised traditional approach.

The results show also that there is an
inverse reciprocity between the handover rate and the measured MAE. 
The lowest MAE is recorded for a handover probability of $h=1$.
This can be intuitively explained by 
the observation that 
since each vehicle requests service as long as it arrives at the BS,
the total number of calls is equal to or larger than 
the total number of vehicles. In all other cases, some of the
vehicles might travel through the cell without  being involved in a call %making a call,
or while making several calls. Then, the estimated number of vehicles
does not provide a lower bound on the number of calls.

The results also suggest that the call arrival rate 
does not have any pronounced effect on the prediction error. 
In contrast, 
the range of the cell does impact the MAE: % as for 
cells that cover a larger segment of the highway, 
the prediction error is smaller. 
Similarly to the effect of the handover probability, 
this result might be explained by the fact that longer distances---under 
otherwise identical conditions namely, speed and service request rate---translate into 
longer dwell time in the cell. 
Then, there is a higher probability 
that the vehicles will make a call while traversing the cell. 
Likewise, the vehicular flow 
might %be a good indicator of 
serve as a lower bound on
the number of service requests
in this case too.

%We highlight that it is more challenging to forecast over short than long time horizons 
%because the traffic is not averaged and consequently less smoother in the former case. 
%The results can be  improved by labelling the days of the week and the hours of the day.
%However, in this work we are not interested in absolute values but
%in the relative performance between  Net and Net\&Road cases, a proof of concept.

%
%
%\begin{figure}[!t]
%	\centering
%	\includegraphics[width=2.5in]{Data}
%	\caption{This is the caption for one fig.}
%	\label{fig1}
%\end{figure}

\section{Conclusion}

The main contribution reported in the paper is a novel paradigm through which 
mobile cellular traffic forecasting is made substantially more accurate. 
Specifically, by incorporating freely available road metrics
we characterise the data generation process and spatial dependencies.
Therefore, this provides a means for improving the forecasting  estimates. 
We employ highway flow and average speed variables together with 
a cellular network traffic metric 
in a light learning structure to predict 
the short-term future load on a cell covering a segment of a highway. 
This is in sharp contrast to prior art that mainly studies urban scenarios 
(with pedestrian and limited vehicular speeds) and develops machine learning approaches 
that use exclusively network metrics and meta information to make mid-term and long-term predictions. 
The learning structure can be used at a cell or edge level,
and can find application in both federated and centralised learning.

%In contrast to prior art, we employ a light learning model---a 
%type of a recurrent neural network---to 
%achieve substantial improvements 
%in mobile cellular traffic prediction performance. 
%These improvements are attained by 
%enriching the data set with 
%freely available metrics 
%that characterise the data generation process 
%and spatial dependencies.
%Specifically, the employed metrics are the vehicular flow and its average speed.
%They provide means to count the potential number of users of the mobile system  
%and hence, to improve the forecasting  estimates. 
%The learning structure can be used at a cell or edge level,
%and can find application in both federated and centralised learning.

The presented machine learning approach is an instance of 
a concept with a much broader scope.
Feeding the mobile cellular network with vehicular traffic data and vice versa---the 
intelligent transportation systems with relevant wireless communication feedback---can 
be used in a myriad of possible ways
to mutually optimise the mobile communication and the intelligent transportation systems.
These promising  pathways are high on our present research agenda.

\section*{Acknowledgments}
%This should be a simple paragraph before the References to thank those individuals and %institutions who have supported your work on this article.
%The author thanks Dr. Pat O'Sullivan for fruitful discussions on the patent application, which 
%were beneficial at the writing stage of this paper and to Professor Pauliina Ilmonen 
%for helpful comments that improved the clarity of the presentation.

The author is grateful to Dr. Pat O'Sullivan for fruitful discussions on the patent application
and to Associate Professor Pauliina Ilmonen for constructive comments on an earlier version of this paper.
Both the discussions and comments helped improving the clarity of the presentation.

\newpage


\begin{thebibliography}{1}
\bibliographystyle{IEEEtran}

\bibitem{piRoad} A. Okic, L. Zanzi, V. Sciancalepore, A. Redondi and X. Costa-Pérez, ``$\pi$-ROAD: a learn-as-you-go framework for on-demand emergency slices in V2X scenarios," 
in \textit{Proc. IEEE Conference on Computer Communications, IEEE INFOCOM 2021}, pp. 1-10, doi: 10.1109/INFOCOM42981.2021.9488677.

\bibitem{ho} Y. Fang, S. Ergüt and P. Patras, ``SDGNet: a handover-aware spatiotemporal graph neural network for mobile traffic forecasting,"  
\textit{IEEE Communications Letters}, vol. 26, no. 3, pp. 582-586, March 2022, doi: 10.1109/LCOMM.2022.3141238.

\bibitem{crossDomain} C. Zhang, H. Zhang, J. Qiao, D. Yuan, and M. Zhang, ``Deep transfer learning for intelligent cellular traffic prediction based on cross-domain big data," 
\textit{IEEE Journal on Selected Areas in Communications}, vol. 37, no. 6, 2019, 1389-1401.

\bibitem{stTran} Q. Liu, J. Li and Z. Lu, ``ST-Tran: spatial-temporal transformer for cellular traffic prediction," 
 \textit{IEEE Communications Letters}, vol. 25, no. 10, pp. 3325-3329, Oct. 2021, doi: 10.1109/LCOMM.2021.3098557.

\bibitem{eff} N. Zhao, A. Wu, Y. Pei, Y.-C. Liang and D. Niyato,``Spatial-temporal aggregation graph convolution network for efficient mobile cellular traffic prediction,"
\textit{IEEE Communications Letters}, vol. 26, no. 3, pp. 587-591, March 2022, doi: 10.1109/LCOMM.2021.3138075.

\bibitem{agg} H. Shi, C. Pan,  L. Yang, and X. Gu, ``AGG: A novel intelligent network traffic prediction method based on joint attention and GCN-GRU,"  
\textit{Wiley Hindawi Security and Communication Networks}, vol. 2021, pp. 1--11.

\bibitem{IoT} L. Fang, X. Cheng, H. Wang and L. Yang, ``Mobile demand forecasting via deep graph-sequence spatiotemporal modeling in cellular networks," 
\textit{IEEE Internet of Things Journal}, vol. 5, no. 4, pp. 3091-3101, Aug. 2018, doi: 10.1109/JIOT.2018.2832071.

\bibitem{citywide} C. Zhang, H. Zhang, D. Yuan and M. Zhang, ``Citywide cellular traffic prediction based on densely connected convolutional neural networks," \textit{IEEE Communications Letters}, vol. 22, no. 8, pp. 1656-1659, Aug. 2018, doi: 10.1109/LCOMM.2018.2841832.


\bibitem{autoMEC} U. Fattore, M. Liebsch, B. Brik and A. Ksentini, ``AutoMEC: LSTM-based user mobility prediction for service management in distributed MEC resources," 
in \textit{Proc. 23rd ACM Int. Conf. Modeling, Analysis and Simulation Wireless Mobile Systems (MSWIM)}, pp. 155-159, Nov. 2020.


\bibitem{campolo} C. Campolo, A. Molinaro, A. Iera and F. Menichella, ``5G Network Slicing for Vehicle-to-Everything Services,"  
\textit{IEEE Wireless Communications}, vol. 24, no. 6, pp. 38-45, Dec. 2017, doi: 10.1109/MWC.2017.1600408.


\bibitem{bigData} J. Wang et al., ``Spatiotemporal modelling and prediction in cellular networks: A big data enabled deep learning approach," 
in \textit{Proc. IEEE Conference on Computer Communications, IEEE INFOCOM 2017}, pp. 1-9, doi: 10.1109/INFOCOM.2017.8057090.


\bibitem{pems} C. Chen, K. Petty, A. Skabardonis, P. Varaiya, and Z. Jia, ``Freeway performance measurement system: mining loop detector data," 
\textit{Transportation Research Record}, vol. 1748, no. 1, pp. 96-102, 2001.

\bibitem{LSTM} S. Hochreiter and J. Schmidhuber, ``Long short-term memory," 
\textit{Neural computation}, vol. 9, no. 8, 1997, pp. 1735-1780.





\end{thebibliography}
\end{document}